\documentclass[runningheads]{llncs}
\usepackage{graphicx}
\usepackage{comment}
\usepackage{amsmath,amssymb} 
\usepackage{color}
\usepackage[ruled]{algorithm2e}
\usepackage{multirow}
\usepackage{threeparttable}
\usepackage{booktabs}

\begin{document}
\pagestyle{headings}
\mainmatter
\def\ECCVSubNumber{3853}  

\title{Learning Connectivity of Neural Networks from a Topological Perspective}


\titlerunning{Learning Connectivity of Neural Networks from a Topological Perspective}

\author{Kun Yuan\orcidID{0000-0002-3681-2196} \and
Quanquan Li\orcidID{0000-0002-8518-2346} \and \\
Jing Shao \and
Junjie Yan}
\authorrunning{Kun Yuan et al.}
%
\institute{SenseTime Research Institute\\
\email{\{yuankun,liquanquan,shaojing,yanjunjie\}@sensetime.com}}
\maketitle

\begin{abstract}

Seeking effective neural networks is a critical and practical field in deep learning.
Besides designing the depth, type of convolution, normalization, and nonlinearities,
the topological connectivity of neural networks is also important.
Previous principles of rule-based modular design simplify the difficulty of building
an effective architecture, but constrain the possible topologies in limited spaces.
In this paper, we attempt to optimize the connectivity in neural networks.
We propose a topological perspective to represent a network into a complete graph for analysis, 
where nodes carry out aggregation and transformation of features, 
and edges determine the flow of information.
By assigning learnable parameters to the edges which reflect the magnitude of connections,
the learning process can be performed in a differentiable manner.
We further attach auxiliary sparsity constraint to the distribution of connectedness, 
which promotes the learned topology focus on critical connections.
This learning process is compatible with existing networks and owns adaptability
to larger search spaces and different tasks.
Quantitative results of experiments reflect the learned connectivity is superior to 
traditional rule-based ones, such as random, residual and complete.
In addition, it obtains significant improvements in image classification and 
object detection without introducing excessive computation burden.

\end{abstract}

\section{Introduction}

Deep learning successfully transits the feature engineering 
from manual to automatic design.
It marks the mapping function from sample to feature 
can be optimized accordingly.
As a tendency, seeking effective neural networks gradually 
becomes an important and practical direction.
But the design of architecture is still a challenging and time-consuming effort.
Part of the research focuses on how depth 
\cite{krizhevsky2012imagenet,glorot2011deep,krizhevsky2012imagenet,perez2018efficient}, 
type of convolution \cite{dai2017deformable,howard2017mobilenets}, 
normalization \cite{wu2018group,luo2018differentiable}
and nonlinearities \cite{maas2013rectifier,ramachandran2017searching} 
affect the performance.
In addition to these endeavors, another group of work also attempted to simplify
the architecture design through stacking blocks/modules and wiring topological connections. 

This strategy was demonstrably first popularized by the VGGNet \cite{simonyan2014very}
that is directly stacked by a series of convolution layers with plain topology.
Due to the problems of gradient vanishing and exploding, extending the network to a deeper
level for better representation is nearly difficult.
To better adapt the optimization process of gradient descent, 
GoogleNet \cite{szegedy2015going} adopted parallel modules, 
and Highway networks \cite{srivastava2015highway} utilized gating units to regulate the flow of information,
resulting in elastic topologies.
Driven by the significance of depth, the residual block consisted of residual mapping
and shortcut was raised in ResNet \cite{he2016deep}.
Topological changes in neural networks successfully scaled up neural networks to 
hundreds or even thousands of layers.
The proposed residual connectivity was widely approved and applied in the following works, e.g.
MobileNet \cite{sandler2018mobilenetv2,howard2019searching} and ShuffleNet \cite{zhang2018shufflenet}.
Divergent from aforementioned relative sparse topologies, 
DenseNet \cite{huang2017densely} wired densely among blocks to reuse features fully.
Recent advances in computer vision also explore neural architecture search (NAS) 
methods \cite{zoph2018learning,liu2018darts,tan2019mnasnet} to search convolutional blocks.
To trade-off efficiency and performance, most of them used hand-designed stacked patterns,
and constrained the search space in limited ones.
These trends reflect the great impact of topology on the optimization of neural networks.
To a certain degree, previous principles of modular design simplify the difficulty of building
an effective architecture.
But how to aggregate and distribute these blocks is still an open question.
Echoing this perspective, we wonder: 
can connectivity in neural networks be learned? What is the suitable route to do this?

\begin{figure}[t!]
  \label{top_res}
  \begin{center}
  \includegraphics[width=0.9\linewidth]{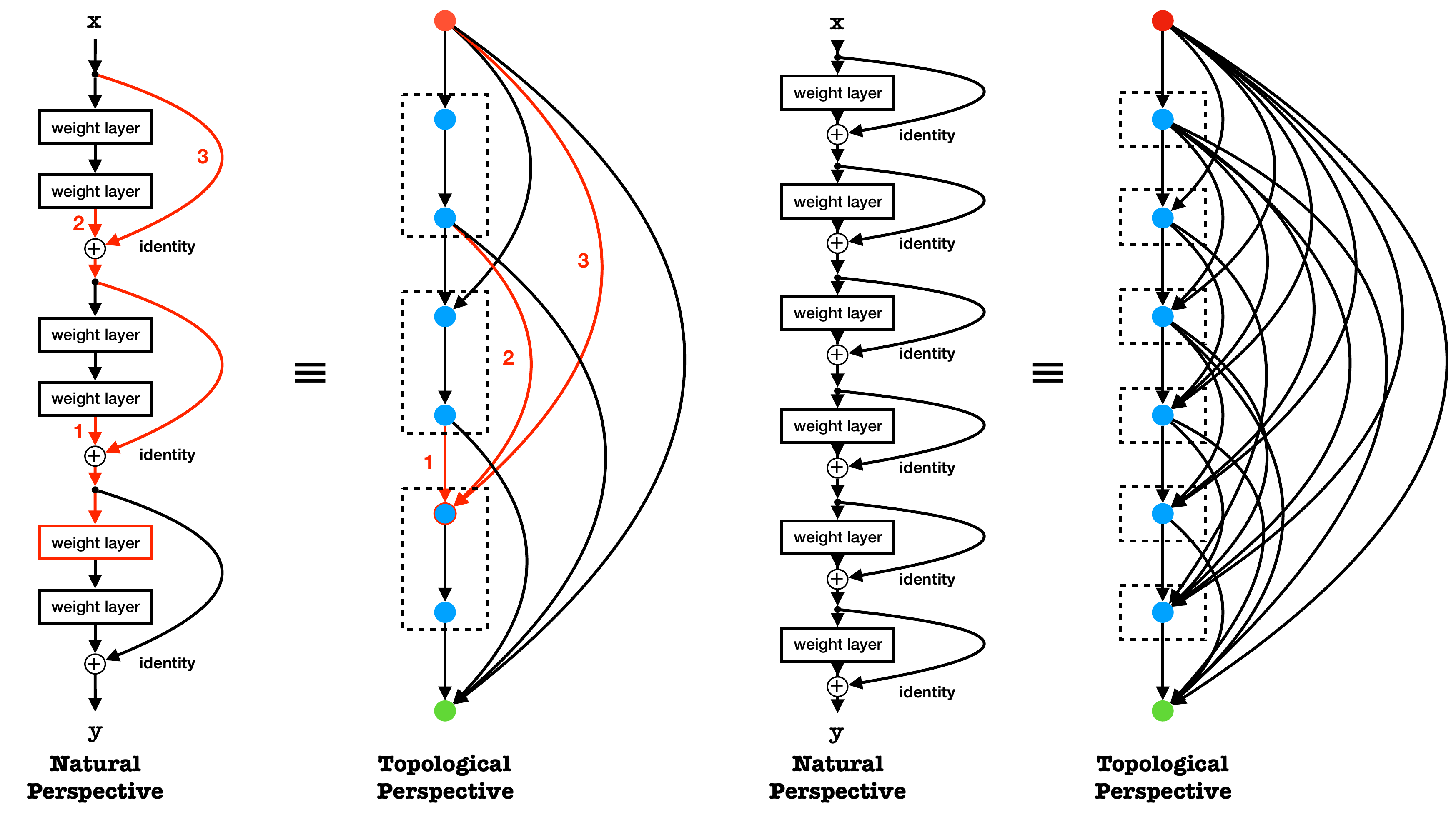}
  \end{center}
  \caption{
     From a natural perspective to the topological perspective 
     for networks with residual connectivity.
     Two types of networks with $1/2$ interval are given.
     Red node denotes the input $\mathbf{x}$, and green one means the output feature $\mathbf{y}$.
     Red arrows give an example of this mapping for a node with in-degree of 3.
     }
  \label{fig:view}
\end{figure}

To answer these questions, we propose a topological perspective to represent
neural networks, resulting in a directed acyclic graph as shown in Fig. \ref{fig:view}.
Under this perspective, transformations (e.g convolution, normalization and 
activation) are mapped into a node,
and connections between layers are projected to edges which indicate the flow of information.
We \textit{first} unfold the residual connections to be a complete graph.
This gives another way to explain the effectiveness of the residual topology,
and inspires us to define the search space using a complete graph.
Instead of choosing a predefined rule-based topology, we assign learnable 
parameters which determine the importance of corresponding connections to the edges.
To adequately promote generalization and concentrate on critical connections,
we attach auxiliary sparsity constraint on the weights of edges.
Particularly, we propose two updating methods to optimize the weights of topology.
One is a {\it uniform} type that regulates different edges uniformly.
The other is an {\it adaptive} type that is logarithmically related to the in-degree
of a node.
Then the connectivity is learned simultaneously with the weights of the network 
by optimizing the loss function according to the task using a modified version of gradient descent.

We evaluate our optimization method on classical networks, such as ResNets and MobileNet.
It demonstrates the compatibility with existing networks and adaptability to larger search spaces.
To exhibit the benefits of connectivity learning, 
we construct a larger search space in which different topologies can be compared strictly.
We also evaluate our method on various tasks and dataset, concretely, 
image classification on CIFAR-100 and ImageNet, object detection on COCO.
Our contributions are as follows:
\begin{itemize}
   \item The proposed topological perspective can be used to represent most existing neural
         networks. For the residual topology, 
         we reveal for the first time the properties of its dense connections,
         which can be used for the search space.
   \item The proposed optimization method is compatible with existing networks.
         Without introducing much additional computing burden, we achieve $2.23\%$
         improvement using ResNet-110 on CIFAR-100, and $0.75\%$ using deepened MobileNet on ImageNet.
   \item We design an architecture called TopoNet for larger search spaces and restrict comparison.
         Quantitative results prove the learned connectivity is superior to random, residual and complete ones, 
         and surpasses ResNet in the similar computation cost by $2.10\%$ on ImageNet.
   \item This method owns good generalization.
         The optimized topology with learned connectivity surpasses the best rule-based one by $0.95\%$ in AP
         on COCO. To the equal-sized backbone of ResNet, the improvement is $5.27\%$.
         We also explore the properties of the optimized topology for future work.
 \end{itemize}

\section{Related Work}

We briefly review related works in the aspects of neural network structure design and 
relevant optimization methods.

Neural network design is widely studied in previous literature.
From shallow to deep, the shortcut connection plays an important role.
Before ResNet, an early practice \cite{venables2013modern} also added linear layer connected 
from input to the output to train multi-layer perceptrons.
Besides, ``Inception'' layer was proposed in \cite{szegedy2015going} 
that is composed of a shortcut branch and a few deeper branches.
Except on large networks, shortcut also proved effective in small networks,
e.g. MobileNet \cite{sandler2018mobilenetv2}, ShuffleNet \cite{zhang2018shufflenet} 
and MnasNet \cite{tan2019mnasnet}.
The existence of shortcut eases vanishing/exploding gradients \cite{he2016deep,srivastava2015highway}.
In this paper, we explain from a {\it topological perspective} that
shortcuts offer dense connections and benefit optimization.
On the macrostructure, there also exist many networks with dense connections.
DenseNet \cite{huang2017densely} contacted all preceding layers and passed
on the feature maps to all subsequent layers in a block.
HRNet \cite{sun2019high} benefits from dense high-to-low connections for fine representations.
Densely connected networks promote the specific task of localization \cite{tang2018quantized}.
Differently, we optimize the desired network from the complete graph in a differentiable way.
And it is different from MaskConnect \cite{ahmed2018maskconnect} which is constrained by $K$ discrete in-degree
and owns binary connections.
This also provides an extension to \cite{xie2019exploring} 
where random graphs generated by different generators are employed to form a network.

For the learning process, our method is consistent with DARTS \cite{liu2018darts} which is differentiable.
In contrast to DARTS, we do not adopt alternative optimization strategies for weights and architecture.
Joint training can replace the transferring step from one task to another, and obtain task-related topology.
Different from sample-based optimization methods \cite{real2019regularized},
the connectivity is learned simultaneously with the weights of the network using our
modified version of gradient descent.
\cite{bender2018understanding,guo2019single} also explored this type and utilized weight-sharing 
across models to amortize the cost of training.
Searching from the full space is evaluated in object detection by NAS-FPN \cite{ghiasi2019fpn},
in which the feature pyramid is sought in all cross-scale connections.
In the aspect of semantic segmentation, Auto-DeepLab \cite{liu2019auto}
formed a hierarchical architecture to enlarge search spaces.
The sparsity constraint can be observed in other applications,
e.g. path selection for a multibranch network \cite{huang2018data},
and pruning unimportant channels for fast inference \cite{han2015learning}.

\section{Methodology}

\subsection{Topological Perspective of Neural Networks}

\begin{figure}[t!]
  \label{top_res}
  \begin{center}
  \includegraphics[width=0.8\linewidth]{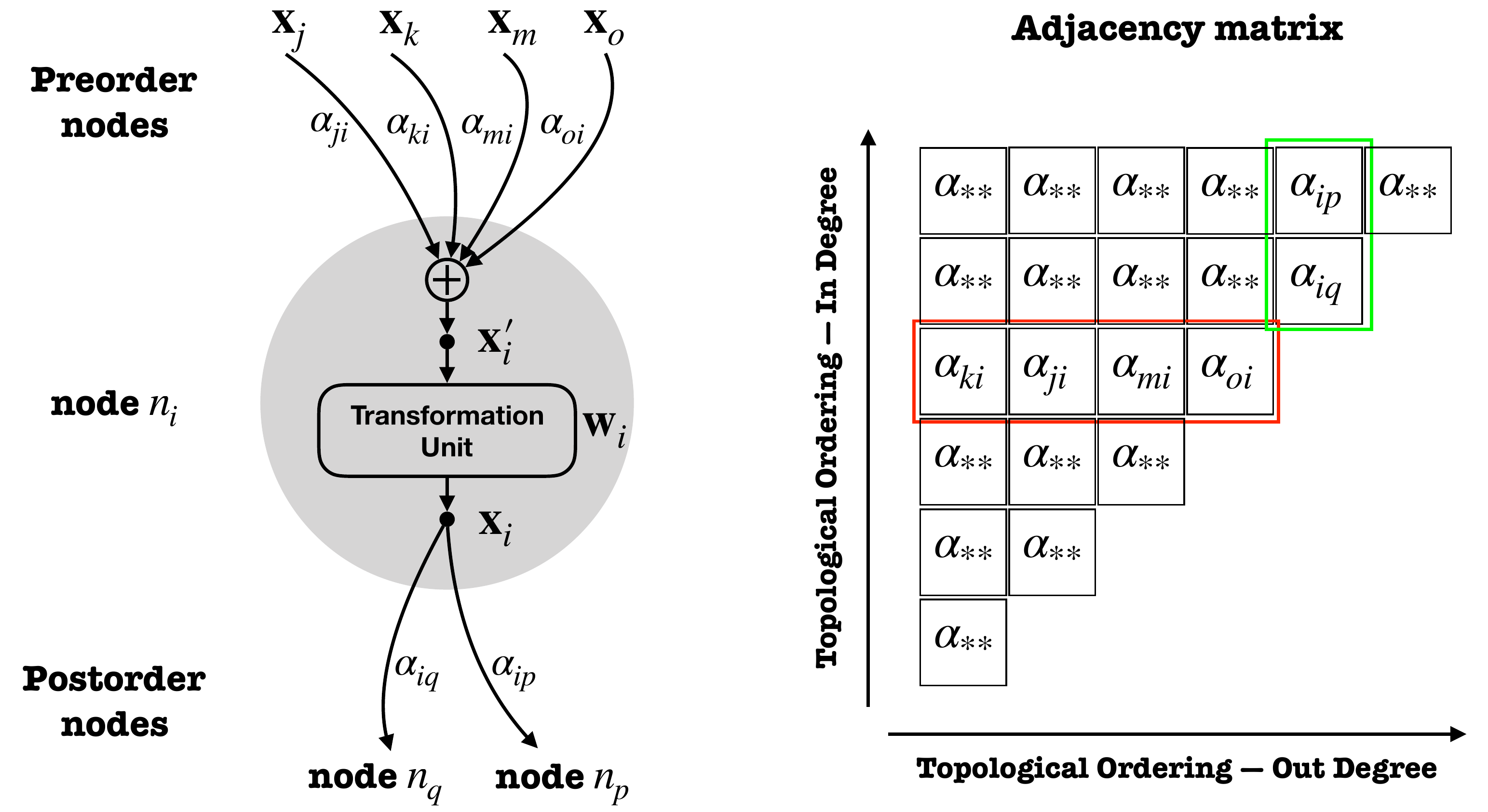}
  \end{center}
  \caption{
     Details of node operations and the adjacency matrix.
     For each node, features generated by preorder nodes are aggregated through 
     weights of edges.
     Then a transformation unit which consists of convolutional layers, batch normalization
     and the activation function is used to transform features.
     Next, features are allocated to postorder nodes where connections exist.
     For a stage, weights of edges can be represented in an adjacency matrix,
     in which rows denote the weights of input edges, and columns stand for 
     the weights of output edges.
     }
  \label{fig:node}
\end{figure}

We represent the neural network using a directed acyclic graph (DAG) in topology.
Specifically, we map both combining (e.g., addition) and transformation 
(e.g., convolution, normalization and activation) to a node.
And connections between layers are represented as edges, which determine the flow of information.
Then we can get a new representation of the architecture $\mathcal G=(\mathcal N, \mathcal E)$,
where $\mathcal N$ is the set of nodes, and $\mathcal E$ denotes the set of edges.

In the graph, each node $n_i \in \mathcal N$ performs a transformation operation $o_i$,
parametrized by $\mathbf{w}_i$, where $i$ stands for the topological ordering of the node.
While the edge $e_{ji} = (j,i,\alpha_{ji}) \in \mathcal E $ means the flow of features
from node $j$ to node $i$, and the importance of the connection is determined by 
the weight of $\alpha_{ji}$.
During forward computation, each node aggregates inputs from preorder nodes where
connections exist.
Then it performs a feature transformation to get an output tensor $\mathbf{x}_i$.
And $\mathbf{x}_i$ is sent out to the postorder nodes through the output edges.
It can be seen in the left of Fig. \ref{fig:node}.
It can be formulated as follows:
\begin{equation}
  \mathbf{x}_i=o_i(\mathbf{x}_i^\prime; \mathbf{w}_i), \ \mbox{where} \ 
  \mathbf{x}_i^\prime = \textstyle\sum\limits_{(j<i) \wedge (e_{ji}\in \mathcal E)} \
  \alpha_{ji} \cdot \mathbf{x}_j.
  \label{eql:node}
\end{equation}

In each graph, the first node in topological ordering is the input one, 
which only performs the distribution of features.
The last node is the output one, which only generates final output of the graph by
gathering preorder inputs.
We also propose an adjacency matrix as the memory space to store weights of edges.
As shown in the right of Fig. \ref{fig:node}, each row denotes the weights of input edges,
and each column is the outputs.
For nodes where there are no edges attached, the corresponding $\alpha$ is $0$.
The dimension of row(with $\alpha \neq 0$) is called in-degree for a node, 
and the dimension of column(with $\alpha \neq 0$) is named as out-degree.

For a network with $k$ stages, $k$ DAGs are initialized and connected in series.
Each graph is linked to its preceding or succeeding stage by output or input node.
We rewrite the weights of nodes as $\mathbf{w}_i^k$ and the weights of edges as $\alpha_{ji}^k$.
For the $k$-th stage, $\mathcal{T}^k(\cdot)$ denotes the mapping function established by
$\mathcal{G}^k$ with parameters of $\mathbf{W}^k$ and $\boldsymbol{\alpha}^k$, 
where $\mathbf{W}^k$ is the set of $\{\mathbf{w}_{i}^{k}\}$, $\boldsymbol{\alpha}^k$
is the set of $\{\alpha_{ji}^{k}\}$.
Given an input $\mathbf{x}$ and corresponding label $\mathbf{y}$, the mapping function from the sample to the feature
representation can be written as:
\begin{equation}
   \mathcal{F}(\mathbf{x}) = \mathcal{T}^k(\cdots \mathcal{T}^2
   (\mathcal{T}^1(\mathbf{x}; \boldsymbol{\alpha}^1, \mathbf{W}^1); 
   \boldsymbol{\alpha}^2, \mathbf{W}^2) \cdots; \boldsymbol{\alpha}^k, \mathbf{W}^k),
\end{equation}

\subsection{Search Space}

By defining the topological perspective of neural networks, most previous networks
can be reformulated from the natural perspective.
For definiteness and without loss of generality, 
we selected the widely used residual connections for analysis 
\cite{he2016deep,sandler2018mobilenetv2,zhang2018shufflenet}.
A block with residual connection formulates $x+\varphi(x)$ as a basic component, 
in which $x$ represents the identity shortcut,
and $\varphi(x)$ denotes the residual mapping.
Normally, the residual component is composed of several repeated weighted layers.
We call the number of repeats as {\it interval}, which is noted as $l$.
Fig. \ref{fig:view} presents two residual architectures with a interval of $1$ and $2$ respectively.
By using Eql.(\ref{eql:node}), we map the architecture from the natural perspective
to the topological perspective.
We give an example of this mapping in red lines.
From a natural perspective, the layer acquires information through skip connections.
In the new topological perspective, the node obtains information by corresponding edges.
It should be pointed out that these two perspectives are completely equivalent in results.
It also can be seen that the residual connections are rather denser than the original view,
and perform multiple feed-forward paths instead of a single deep network.
Our topological view explains the reason why residual connectivity is effective from a new aspect
different from \cite{DBLP:conf/nips/VeitWB16}.

If the interval degrades to $1$, as shown in the right of Fig. \ref{fig:node}, 
its topology evolves into a complete graph.
Structurally, all nodes are directly connected to the input and output,
resulting in indirect access to the gradients and the original input.
Different from stacking blocks using predefined connectivity,
the complete graph provides all possible connections 
and is suitable to be the search space.
For a complete graph with $N$ nodes, the search space contains $2^{N(N-1)/2}$ possible topological structures.
For a network with $k$ stages, the total search space can be noted as $\prod\nolimits_{k} 2^{N^k(N^k-1)/2}$.
And it is much wider than cell-based or block-based approaches 
\cite{tan2019mnasnet,guo2019single,howard2019searching}.
By assigning learnable parameters which reflect the 
magnitude of connections to edges, it changes to a weighted graph.
Within the search space, the connectivity can be optimized by learning continuous
weights of edges.



\subsection{Optimization of Topological Connectivity}

We put forward a differentiable type to optimize the topological connectivity
by learning a set of continuous weights of edges $\boldsymbol{\alpha}$.
And they are learned simultaneously with all the other weights in the network via the loss
generated by the concurrent task, noted as $\mathcal{L}_{t}(\cdot)$.
Different from \cite{ahmed2018maskconnect}, we do not transform the weights of $\boldsymbol{\alpha}$ into binary.
This allows us to assign discriminating weights to different feature inputs.
Different from the selection of node type \cite{liu2018darts},
we do not select the maximum input edge using $\arg\max$ operation.
Instead, the continuous weights guarantees the consistency between training and testing.
The optimization objective can be viewed as:
\begin{equation}
   \min_{\mathbf{W}, \boldsymbol{\alpha}} \mathcal{L}_{t}
   (\mathcal{F}(\mathbf{x}; \mathbf{W}, \boldsymbol{\alpha}), \mathbf{y})
\end{equation}
Set $\frac{\partial \mathcal{L}_{t}}{\partial \mathbf{w}_i}$ be the gradients that
the network flows backwards to $\mathbf{w}_i$.
And let $\frac{\partial \mathcal{L}_{t}}{\partial \mathbf{x}_i}$ be the gradients to $\mathbf{x}_i$.
Then the gradients update to $\mathbf{w}_i$ and $\alpha_{ji}$ are of the form:
\begin{eqnarray}
  \mathbf{w}_i & \leftarrow & \mathbf{w}_i + \eta \frac{\partial \mathcal{L}_{t}}{\partial \mathbf{w}_i} \label{eql:w}\\
  \alpha_{ji}  & \leftarrow & \alpha_{ji} + \eta \sum \frac{\partial \mathcal{L}_{t}}{\partial \mathbf{x}_i} \odot
  \frac{\partial o_i}{\partial \mathbf{x}_i^\prime} \odot \mathbf{x}_j,
  \label{eql:grad}
\end{eqnarray}
where $\eta$ is the learning rate, and $\odot$ indicates entrywise product.


Since the features generated by different layers exhibit different semantic representations,
they contribute differently to subsequent layers, resulting in diversities of the importance
of connections.
Much as the mammalian brain \cite{rauschecker1984neuronal} in biology,
where synapses are created in the first few months of a child's development,
followed by gradual re-weighting through postnatal knowledge,
growing into a typical adult with relative sparse connections.

\begin{algorithm}[t!]\label{algorithm}
\caption{Optimization of Topological Connectivity.}
\small
\SetKwInOut{Input}{Input}  
\Input{Network represented with $k$ graphs $\mathcal{G}(\mathcal N, \mathcal E; \mathbf{W}, \boldsymbol{\alpha})$,
       training set $\{(\mathbf{x}$, $\mathbf{y})^{(s)}\}$,
       {\it Type} of sparsity, $\lambda$ and $\eta$.}
\SetKwInOut{Output}{Output} 
\Output{Optimized weights $\mathbf{W}$ and a weighted graph defined by topology $\boldsymbol{\alpha}$.}
\For{$(\mathbf{x}^s, \mathbf{y}^s)$ in training set}{
  \For{$\mathcal{G}^k$ in network}{
    Set $rev\_adj\_dict = \{\}$ to save directed edges\;
    Set $memory=\{\}$ to save features generated by nodes\;
    \For{$e_{ji}$ in $\mathcal{E}^k$}{
      $rev\_adj\_dict[$i$].append(j)$\;
    }
    \For{$n_i$ in $\mathcal{N}^k$}{
      Obtain indices of input nodes of $n_i$ from $rev\_adj\_dict$[$i$] and 
      fetch corresponding input features from $memory$\;
      Aggregate inputs by weights of $\boldsymbol{\alpha}^k$ and transform features using Eql.(\ref{eql:node}) and 
      store transformed features $memory$[i]$=\mathbf{x}_i$\;
    }
    Fetch the features generated by the output node for $\mathcal{G}^{k+1}$\; 
  }
  Obtain final representation and compute the loss w.r.t $\mathbf{y}^s$ by Eql.(\ref{eql:loss})\;
  Compute gradients w.r.t $\mathbf{w_i}$ in $\mathbf{W}$ by Eql.(\ref{eql:w}) and update weights\;
  \uIf{{\it Type} \textbf{is} {\it uniform}}{
    Compute gradients w.r.t $\alpha_{ji}$ in $\boldsymbol{\alpha}$ 
    by Eql.(\ref{eql:uni}) and update weights\;
  }
  \uElseIf{{\it Type} \textbf{is} {\it adaptive}}{
    Compute gradients w.r.t $\alpha_{ji}$ in $\boldsymbol{\alpha}$ 
    by Eql.(\ref{eql:ada}) and update weights\;
  }
}
\end{algorithm}

To facilitate this process appropriately, we raise to attach sparsity constraint as a regularization
on the distribution of weights of edges.
Similar thought also has been verified in hashing representation \cite{weinberger2009feature}
that sparsity can bring effective gain through minimizing a hash collision.
We choose L1 regularization, denoted as $\mathcal{L}_{1}(\cdot)$, 
to penalize non-zero parameters of edges resulting in more parameters near zero.
This sparsity constraint promotes attention to more critical connections.
Then the loss function of our proposed method can be reformulated as:
\begin{equation}
  \mathcal{L} = \mathcal{L}_{t} + \lambda \mathcal{L}_{1} =
  \mathcal{L}_{t}(\mathcal{F}(\mathbf{x}; \mathbf{W}, \boldsymbol{\alpha}), 
  \mathbf{y}) + \lambda \cdot \|\boldsymbol{\alpha}\|_1,
\label{eql:loss}
\end{equation}
and $\lambda$ is a hyper-parameter to balance the sparse level.
Due to the properties of a complete graph, we propose two types to update $\boldsymbol{\alpha}_{ji}$.
The first one is {\it uniform sparsity} that attaches constraint on all 
weights of edges uniformly.
And let $\frac{\partial \mathcal{L}_{1}}{\partial \mathbf{x}_i}$ be the gradients 
to $\mathbf{x}_i$, we rewrite the Eql.(\ref{eql:grad}) as:
\begin{equation}
  \alpha_{ji} \leftarrow \alpha_{ji} + \eta \sum 
  (\frac{\partial \mathcal{L}_{t}}{\partial \mathbf{x}_i} + \lambda
   \frac{\partial \mathcal{L}_{1}}{\partial \mathbf{x}_i}) \odot
  \frac{\partial o_i}{\partial \mathbf{x}_i^\prime} \odot \mathbf{x}_j,
\label{eql:uni}
\end{equation}
The second one is {\it adaptive sparsity} which is logarithmically 
related to the in-degree $\delta_i$ of a node $n_i$.
It performs larger constraints on dense input and smaller on sparse input.
For the nodes with fewer input edges, this can ensure the smooth flow of information 
and avoid being blocked.
In this type, the $\alpha_{ji}$ is updated by:
\begin{equation}
  \alpha_{ji} \leftarrow \alpha_{ji} + \eta \sum 
  (\frac{\partial \mathcal{L}_{t}}{\partial \mathbf{x}_i} + \lambda \log(\delta_i)
   \frac{\partial \mathcal{L}_{1}}{\partial \mathbf{x}_i}) \odot
  \frac{\partial o_i}{\partial \mathbf{x}_i^\prime} \odot \mathbf{x}_j.
\label{eql:ada}
\end{equation}
These two types will be further discussed in the experiments section.
Algorithm \ref{algorithm} summarizes the optimization procedure detailedly.

\section{Experiments and Analysis}

\subsection{Connectivity Optimization for Classical Networks}\label{optim_classic}

Our optimization method is compatible with classical networks.
To investigate the applicability, we select ResNet-CIFAR \cite{he2016deep} 
consisted of $3\times 3$ \texttt{conv} and MobileNetV2-1.0 
\cite{sandler2018mobilenetv2} consisted of \texttt{Inverted Bottleneck}.
For the optimization of ResNets, we rewire the interval of $2$ 
in the \texttt{BasicBlock} to $1$ to form the complete graph.
For MobileNetV2-1.0, each node involves a residual connection 
and can be viewed as a complete graph naturally.
In the case that MobileNet owns fewer layers in each stage,
we also increase the depth by increasing the node in each stage,
resulting in larger search spaces.
It is also a common skill to expand networks \cite{tan2019efficientnet}.
Through assigning learnable parameters to their edges,
the topologies can be optimized using Algorithm \ref{algorithm}.
It should be mentioned that the additional computations and parameters 
introduced by the edges are negligible compared with convolution.

\begin{table}[h]
   \small
   \caption{Optimization Top-1 Accuracy of ResNets on CIFAR-100.}
   \centering
   \begin{tabular}{cccccc}
      \toprule
      \textbf{Network} & \texttt{Params(M)} & \texttt{FLOPs(G)} & \textbf{Original} & \textbf{Optimized} & \textbf{Gain} \\
      \midrule
      ResNet-20     & 0.28 & 0.04 & 69.01 & $\mathbf{69.91}\pm 0.12$ & $\mathbf{0.90}$ \\
      ResNet-32     & 0.47 & 0.07 & 72.07 & $\mathbf{73.34}\pm 0.09$ & $\mathbf{1.37}$ \\
      ResNet-44     & 0.67 & 0.10 & 73.73 & $\mathbf{75.60}\pm 0.14$ & $\mathbf{1.87}$ \\
      ResNet-56     & 0.86 & 0.13 & 75.22 & $\mathbf{76.90}\pm 0.03$ & $\mathbf{1.68}$ \\
      ResNet-110    & 1.74 & 0.25 & 76.31 & $\mathbf{78.54}\pm 0.15$ & $\mathbf{2.23}$ \\
      \bottomrule
   \end{tabular}
   \label{tab:cifar}
\end{table}

First, we evaluate the optimization of the connectivity 
with ResNets on CIFAR-100 \cite{krizhevsky2009learning}.
The experiments are trained using $2$ GPUs 
\footnote{All of our experiments were performed using NVIDIA Tesla V100 GPUs 
          with our implementation in PyTorch \cite{paszke2017automatic}.} 
with batchsize 128 and weight decay 5e-4.
We follow the hyperparameter settings in paper \cite{devries2017improved}, 
which initializes $\eta = 0.1$ and divides by 5 times at 60th, 120th, 160th epochs.
The training and test size is $32\times 32$.
We report classification accuracy on the validation set by 5 repeat runs.
The results are shown in Table \ref{tab:cifar}.
Under similar \texttt{Params} and \texttt{FLOPs}, 
the optimization brings $2.22\%$ improvement
on Top-1 accuracy for ResNet-110, which reflects larger search spaces
lead to more improvements.

Next, we extend our method to ImageNet dataset \cite{russakovsky2015imagenet} using
MobileNets.
We train MobileNetV2 using $16$ GPUs for 200 epochs with a batch size of 1024.
The initial learning rate is 0.4 and cosine shaped learning rate decay \cite{loshchilov2016sgdr} is adopted.
Following \cite{sandler2018mobilenetv2}, we use a weight decay of 4e-5 and dropout \cite{hinton2012improving} of 0.2.
Nesterov momentum of $0.9$ without dampening is also used.
The training and test size is $224\times 224$.
The network with 2 times of layers is denoted as 2N.
Under the mobile-setting, we achieve $76.4\%$ Top-1 accuracy.
Under the larger optimization space of 6N, the optimization brings a $0.75\%$ improvement.
This further demonstrates the benefits of topology optimization for different networks.

\begin{table}[h]
  \small
  \caption{Optimization Top-1 Accuracy of Scaled MobileNets on ImageNet}
  \centering
  \begin{tabular}{cccccc}
     \toprule
     \textbf{Network} & \texttt{Params(M)} & \texttt{FLOPs(G)} & \textbf{Original} & \textbf{Optimized} & \textbf{Gain} \\
     \midrule
     MobileNetV2-1.0     & 3.51 & 0.30 & 72.60 & $\mathbf{72.86}\pm 0.13$ & $\mathbf{0.24}$ \\
     MobileNetV2-1.0-2N  & 6.43 & 0.60 & 75.93 & $\mathbf{76.40}\pm 0.05$ & $\mathbf{0.47}$ \\
     MobileNetV2-1.0-4N  & 9.62 & 1.10 & 77.33 & $\mathbf{77.87}\pm 0.09$ & $\mathbf{0.54}$ \\
     MobileNetV2-1.0-6N  & 12.00 & 2.06 & 77.61 & $\mathbf{78.36}\pm 0.14$ & $\mathbf{0.75}$ \\
     \bottomrule
  \end{tabular}
  \label{tab:classic}
\end{table}


\vspace{-0.5cm}
\subsection{Expanding to Larger Search Spaces by TopoNet}

Due to restricted optional topologies of classical networks, the topology can be only 
optimized in small search spaces, which limits the representation ability of topology.
These may limit the influence caused by topological changes and affect the search for 
optimal topology.
In this section, we propose a larger search space,
and fully illustrate the improvement brought by topology optimization.
The properties of edges and nodes in the optimized topology are also analyzed.

\begin{table}[h]
  \caption{Architectures of TopoNets for ImageNet.}
  \small
  \centering
  \begin{tabular}{cccccc}
    \toprule
    \multirow{2}{*}{\textbf{Layers}} & \textbf{Output} & \textbf{Component}    & \multicolumn{3}{c}{\textbf{Edges of Different Topologies}} \\
    \cline{4-6}
                            & \textbf{Size}   & \textbf{and Channels} & \texttt{Random}, $p$ & \texttt{Residual}, $l$ & \texttt{Complete} \\
    \midrule
    Head      & $112\times 112$ & $7\times 7$ conv, $C$ & - & - & - \\
    \midrule
    Stage 1   & $56\times 56$ & $N_1$ nodes, $2\times C$   & $p\cdot \tbinom{N_1}{2}$ & $\frac{N_1-2}{l}+\tbinom{\frac{N_1-2}{l}+2}{2}$ & $\tbinom{N_1}{2}$ \\ 
    Stage 2   & $28\times 28$ & $N_2$ nodes, $4\times C$   & $p\cdot \tbinom{N_2}{2}$ & $\frac{N_2-2}{l}+\tbinom{\frac{N_2-2}{l}+2}{2}$ & $\tbinom{N_2}{2}$ \\
    Stage 3   & $14\times 14$ & $N_3$ nodes, $8\times C$   & $p\cdot \tbinom{N_3}{2}$ & $\frac{N_3-2}{l}+\tbinom{\frac{N_3-2}{l}+2}{2}$ & $\tbinom{N_3}{2}$ \\
    Stage 4   & $7\times 7$   & $N_4$ nodes, $16\times C$  & $p\cdot \tbinom{N_4}{2}$ & $\frac{N_4-2}{l}+\tbinom{\frac{N_4-2}{l}+2}{2}$ & $\tbinom{N_4}{2}$ \\
    \midrule 
    Classifier & $1\times 1$  & \texttt{GAP}, 1k-d \texttt{fc}, \texttt{softmax}  & - & - & - \\
    \bottomrule
  \end{tabular}
  \label{tab:arch_detail}
\end{table}

We design a series of architectures named as TopoNets that 
can flexibly adjust search space, types of topology and node.
As shown in Table \ref{tab:arch_detail}, it consists of four stages with number of nodes of $\{N_1, N_2, N_3, N_4\}$.
The topology in each stage is defined by a graph, whose type can be chosen from 
\{\texttt{complete}, \texttt{random}, \texttt{residual}\}.
The \texttt{complete} graph is used for the optimization of topology.
For a more strict comparison, we also take the other two types as baselines.
The \texttt{residual} one is a well-designed topology.
In the \texttt{random} one, an edge between two nodes is linked with probability $p$, 
independent of all other nodes and edges.
The higher the probability, the denser it is.
We follow two simple design rules used in \cite{he2016deep}, 
(\romannumeral1) in each stage, the nodes have the same number of filters $C$;
(\romannumeral2) and if the feature map size is halved, the number of filters is doubled.
The change of filters is implied by the first calculation node in each graph.
For the head of the network, we use a single convolutional layer for simplicity.
The network ends with a classifier composed of a global average pooling (\texttt{GAP}),
a 1000-dimensional \texttt{fully-connected} layer and \texttt{softmax} function.

\subsubsection{Setup for TopoNet.}

To demonstrate the optimization capability in the larger search space and
to compare with existing network, we designed a set with similar computation
cost as ResNet-50.
We select the separable depthwise convolution that includes a $3\times 3$ depthwise convolution 
followed by a $1\times 1$ pointwise convolution,
and build a triplet unit \texttt{ReLU-conv-BN} as the node.
The number of nodes in each stage is \{$14,20,26,14$\}.
In this setting, the number of possible discrete topologies is 
$6\times 10^{209}$.
The weights of $\boldsymbol{\alpha}$ are initialized to be $1$.
And $C$ is set to be $64$, resulting in \texttt{Params} of $23.23M$ and 
\texttt{FLOPs} of $3.95G$ 
(e.g ResNet-50 with \texttt{Params} of $25.57M$ and \texttt{FLOPs} of $4.08G$).

\subsubsection{Strict Comparisons.}

To demonstrate the effectiveness of our optimization method,
we select graphs with \texttt{random}, \texttt{residual} connectivity as baselines.
For comparison, we also reproduce Erd\"{o}s-R\'{e}nyi (\texttt{ER}), Barab\'{a}si-Albert (\texttt{BA}) 
and Watts-Strogatz \texttt{WS} graphs \cite{xie2019exploring} using
\texttt{NetworkX} \footnote{\url{https://networkx.github.io}}.
Since original paper does not release codes, we compare the best configurations of
their method.
We use these graphs to build networks under the same setup in TopoNet.
Two types of sparsity constraints are also demonstrated.
For a fair comparison, all experiments are conducted on ImageNet with training 100 epochs.
We use a weight decay of 1e-4 and a Nesterov momentum of $0.9$ without dampening.
Dropout is not used.
Label smoothing regularization \cite{szegedy2017inception} 
with a coefficient of 0.1 is also used.

\begin{table}[h]
  \caption{Comparision with Different Topologies on ImageNet.}
  \small
  \centering
  \begin{tabular}{ccc}
  \toprule
  \textbf{Network}        & \textbf{Top-1 Acc(\%)} & \textbf{Top-5 Acc(\%)}\\
  \midrule
  ResNet-50       & $76.50$ & $93.10$ \\
  \midrule
  \texttt{random}, $p=0.8$  & $77.56\pm0.22$ & $93.69\pm0.32$\\
  \texttt{random}, $p=0.6$  & $77.84\pm0.19$ & $93.65\pm0.43$\\
  \texttt{random}, $p=0.4$  & $77.90\pm0.27$ & $93.76\pm0.37$\\
  \midrule
  \texttt{residual}, $l=4$  & $77.72\pm0.13$ & $93.57\pm0.20$ \\
  \texttt{residual}, $l=3$  & $78.10\pm0.07$ & $93.83\pm0.17$ \\
  \texttt{residual}, $l=2$  & $78.26\pm0.14$ & $93.78\pm0.22$ \\
  \midrule
  \texttt{ER}, $p=0.2$      & $77.76\pm0.33$ & $93.20\pm0.41$\\
  \texttt{BA}, $m=5$        & $78.08\pm0.17$ & $93.46\pm0.34$\\
  \texttt{WS}, $k=4, p=0.75$& $78.19\pm0.25$ & $93.78\pm0.24$\\
  \midrule
  \texttt{complete}                         & $77.24\pm0.12$ & $93.40\pm0.23$ \\
  \texttt{complete}, $\boldsymbol{\alpha}$  & $78.22\pm0.13$ & $93.80\pm0.15$ \\
  \texttt{complete}, $\boldsymbol{\alpha}$, {\it uniform}    & $\mathbf{78.46}\pm0.14$ & $93.81\pm0.32$ \\
  \texttt{complete}, $\boldsymbol{\alpha}$, {\it adaptive}   & $\mathbf{78.60}\pm0.16$ & $93.92\pm0.11$ \\
  \bottomrule
  \end{tabular}
  \label{tbl:classification}
\end{table}

The validation results are shown in Table \ref{tbl:classification}.
Some conclusions can be drawn from the results.
(\romannumeral1) Topological connectivity of network largely affects the performance of representation.
(\romannumeral2) The performance is related to the density of connections according to
different $p$ and $l$.
(\romannumeral3) For the \texttt{complete} graphs, direct optimization 
with $\boldsymbol{\alpha}$ can yield $0.98\%$ improvement on Top-1.
(\romannumeral4) Through assigning sparsity constraints, performances have been further improved. 
The \texttt{complete} graph with {\it adaptive} sparsity constraint
 gets the best Top-1 of $78.60\%$. This proves the benefits of sparseness for the connectivity.
(\romannumeral5) The connectivity can be optimized in neural networks, and is superior to
rule-based designed ones, such as 
\texttt{random}, \texttt{residual}, \texttt{BA} and \texttt{WS}.

\begin{figure}[t!]
  \begin{center}
  \includegraphics[width=\linewidth]{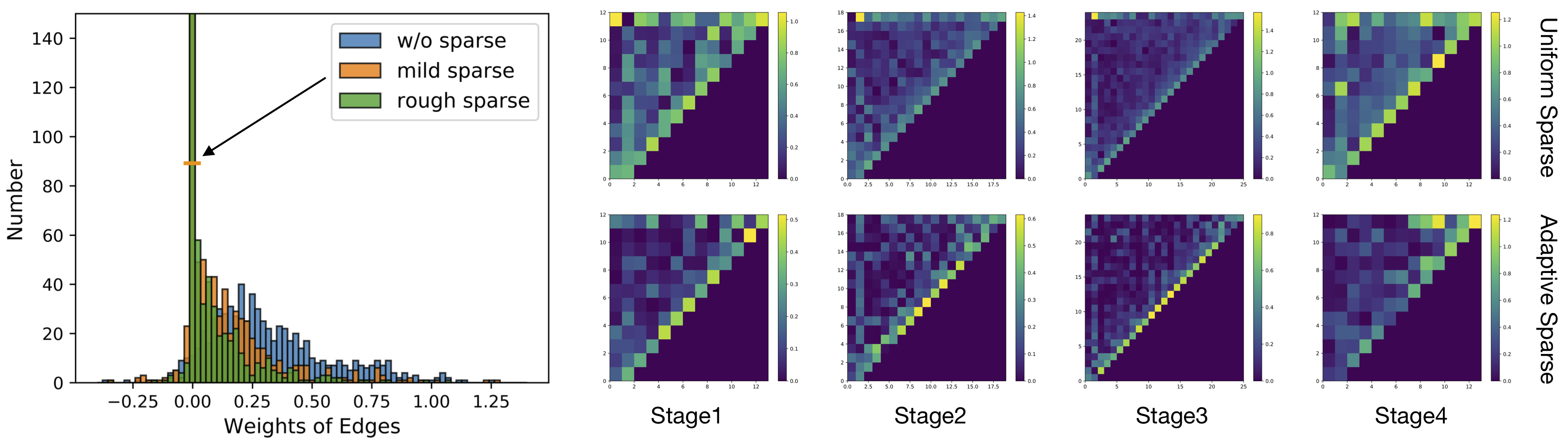}
  \end{center}
  \caption{
    The effect of sparsity constraint on distributions of $\boldsymbol{\alpha}$.
    The histogram on the left indicates that sparsity drives most of the weights near zero. 
    Adjacency matrices on the right shows the difference between {\it uniform} and {\it adaptive} one,
    whose rows correspond to the input edges for a particular node and
    columns represent the output ones.
    Colors indicate the weights of edges.}
  \label{fig:sparse_dist}
\end{figure}

In order to intuitively understand the optimization effect of sparsity constraints on 
dense topological connections, we give the distributions of the learned $\boldsymbol{\alpha}$ Fig. \ref{fig:sparse_dist}.
Sparsity constraints push more parameters near zero, resulting in focusing on critical connections.
With the enhancement of constraints, more connections disappear.
Excessive constraints will damage the representation of features, 
so we set the weight of balance $\lambda$ to be e-4 in all experiments.
And it is also robust in the range from e-5 to e-4, resulting in similar effects.
In the right of Fig. \ref{fig:sparse_dist}, 
we give the optimized results with two types of constraints.
The {\it adaptive} one penalizes denser connections a lot and keeps the 
relative sparser but critical connections, resulting in better performance.

\subsection{Transferability on Different Tasks}

To evaluate the generalization and transferability for both optimization method and TopoNets,
we also conduct experiments on COCO object detection task \cite{lin2014microsoft}.
We adopt FPN \cite{lin2017feature} as the object detection method.
The backbone is replaced with corresponding pretrained one in Table \ref{tbl:classification},
and is fine-tuned on COCO train2017 dataset.
We test using the COCO val2017 dataset.
Our fine-tuning is based on $1\times$ setting of the publicly available 
\texttt{Detectron} \cite{girshick2018detectron}.
The training configurations of different models are consistent.
Test performances are given in Table \ref{tbl:detection}.
To comparable ResNet-50, TopoNets obtain significant promotions in AP with lower 
computation costs.
Contrast with elegant residual topology, our optimization method can also achieve
 increase by $0.95\%$.
These results indicate the effectiveness of the proposed network and the optimization method.

\begin{table}[h]
  \caption{Transferability Results on COCO object detection.}
  \small
  \centering
  \setlength{\tabcolsep}{1.5mm}{
  \begin{tabular}{ccccccc}
     \toprule
     \textbf{Backbone} & \textbf{AP} & \textbf{$\mbox{AP}_{50}$} & 
     \textbf{$\mbox{AP}_{75}$} & \textbf{$\mbox{AP}_{S}$} & \textbf{$\mbox{AP}_{M}$} & \textbf{$\mbox{AP}_{L}$} \\
     \midrule
     ResNet-50  & 36.42 & 58.66 & 38.90 & 21.93 & 39.84 & 46.74 \\
     \midrule
     \texttt{Residual}, $l=2$                                          & 40.74(+4.32) & 63.22 & 44.62 & 25.01 & 44.18 & 52.74 \\
     \texttt{Complete}, $\boldsymbol{\alpha}$                          & 41.35(+4.93) & 63.32 & 45.08   & 25.63 & 44.99 & 53.47 \\
     \texttt{Complete}, $\boldsymbol{\alpha}$, {\it uniform}  & $\mathbf{41.46}$(+5.04) & 63.83 & 44.91 & 25.07 & 45.31 & 53.52 \\
     \texttt{Complete}, $\boldsymbol{\alpha}$, {\it adaptive} & $\mathbf{41.69}$(+5.27) & 63.86 & 45.45 & 25.58 & 45.52 & 53.69 \\
     \bottomrule
  \end{tabular}}
  \label{tbl:detection}
\end{table}

\subsection{Exploring Topological Properties by Graph Damage}

\begin{figure}[h]
\centering
\includegraphics[width=0.9\linewidth]{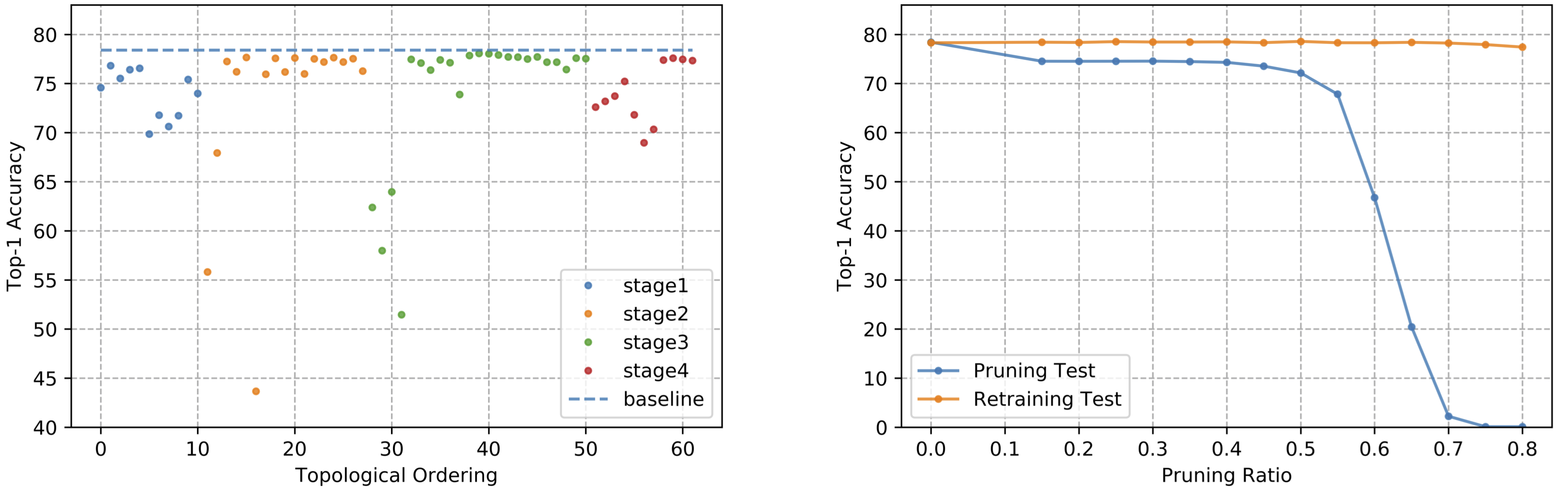}
\caption{Impact of node (left) and edge (right) removal for the optimized topology.}
\label{fig:edge_node}
\end{figure}

We further explore the properties of the optimized topology.
First, we remove individual nodes according to its topological ordering 
in the graph and evaluate it without extra training.
We expect them to break because dropping any layer drastically changes 
the input distribution of all subsequent layers.
Surprisingly, most removals do not lead to a noticeable change as shown 
in Figure \ref{fig:edge_node} (left).
It can be explained that the available paths is reduced from $(n-1)!$ to $(n-2)!$,
leaving sufficient paths.
This suggests that each node in the complete graph do not strongly rely on others,
although it is trained jointly.
Direct links with input/output nodes make each node contribute to the final feature
representation, and benefits the optimization process.
Another observation is that nodes in the front of topological orderings
contribute more.
This can be explained that for a node with the ordering of $i$, the generated $x_i$ can be
only received by node $j$ (where $j>i$).
This causes the feature generated by the front nodes to participate in aggregation
as a downstream input. It makes the front nodes contribute more, 
which can be used to reallocate calculation resources in future work.

Second, we consider the impact of edge removal.
All edges with $\alpha$ below a threshold are pruned from the graph,
only remaining the important connections.
Accuracies before and after retraining are given in Figure \ref{fig:edge_node} (right).
Without retraining, accuracy decreases as the degree of pruning deepens.
It is interesting to see that we have the ``free lunch'' of reducing less than $40\%$
without losing much accuracy.
If we fix $\alpha$ of remaining edges and retrain the weights,
it can maintain accuracy with $80\%$ of the nodes removed.
This proves that the optimization process has found indeed important connections.
After pruning edges, nodes with zero in-degree or zero out-degree maybe
safely removed.
It can be used to reduce the parameters and accelerate inference 
in practical applications.

\subsection{Visualization of the Optimization Process}

\begin{figure}[h]
  \begin{center}
     \includegraphics[width=0.9\linewidth]{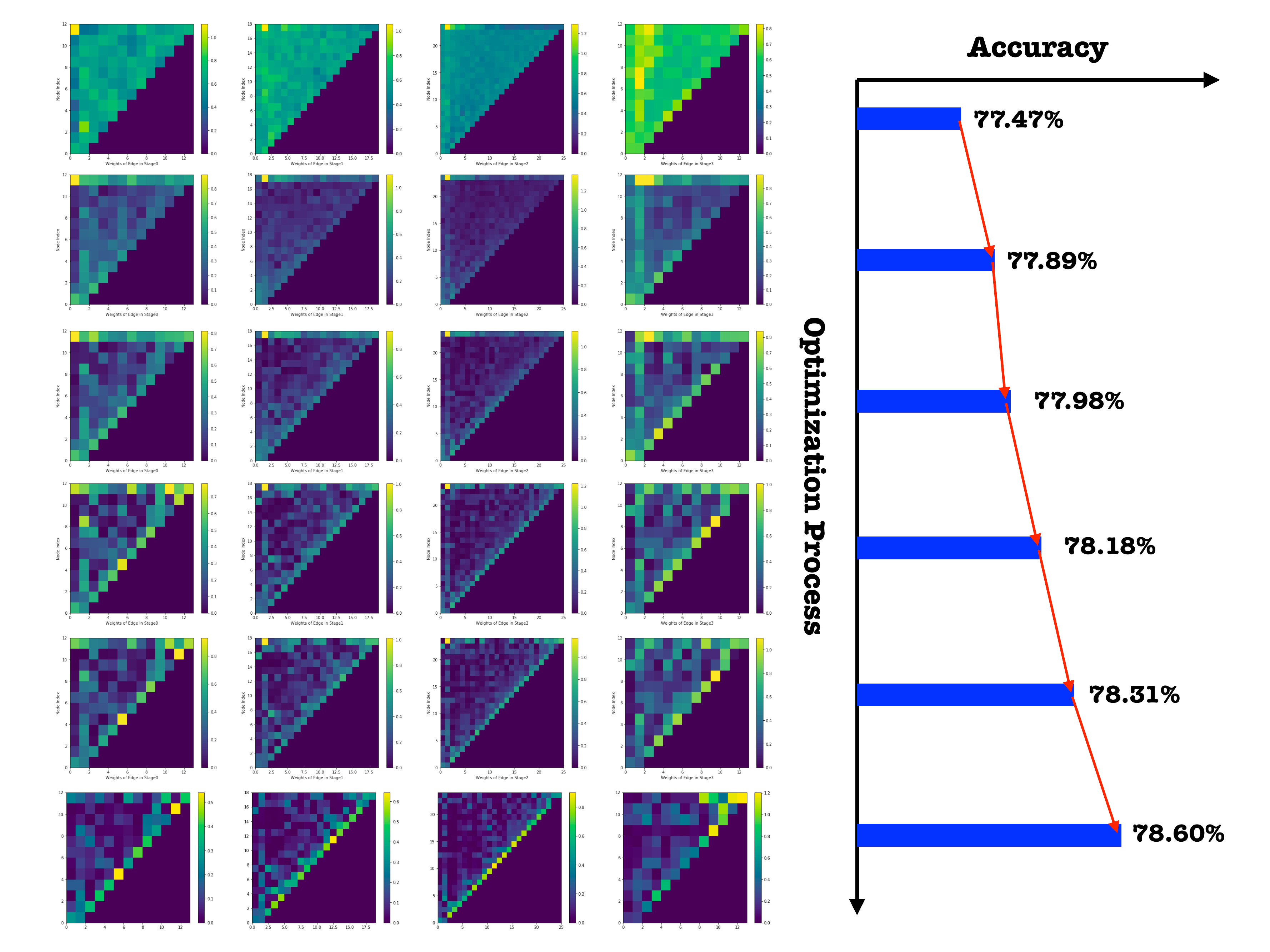}
  \end{center}
  \caption{
     The changes of the connectivity and corresponding accuracies after retraining.
     It proves that the representation capability of topological 
     connectivity improves along with the process,
     and demonstrates the effectiveness of optimization. 
  }
  \label{fig:ckpt}
\end{figure}

We visualize the optimization process in Fig. \ref{fig:ckpt}.
During the initial phase, there are strong connections between all nodes.
As the optimization process progresses, the connections become sparse,
leaving the critical ones.
We sample topologies with different connectivity during the process and retrain them from scratch with $\boldsymbol{\alpha}$ froze.
This allows us to compare the change of topology capabilities during optimization.
Validation accuracies are given in the right of the figure.
It can be seen the representation ability of connectivity
increases with the training process, not just the weights of networks.

\section{Conclusion and Future Work}

In this work, we proposed a feasible way for the learning of topological connectivity
in neural networks.
Motivated by our \textit{topological perspective},
the optimization space is defined as a complete graph.
By assigning learnable continuous weights which reflect the importance of connections,
the optimization process is transformed into a differentiable type with less extra cost.
The sparsity constraint further improve the generalization and performance.
This method is compatible with existing networks, 
and the optimized connectivity is superior to rule-based designed ones.
Experiments on different tasks proved the effectiveness and transferability.
Moreover, the observed properties of topology can be used for future work and
practical applications.
Our work has a wide application and is complementary to existing neural
architecture search methods.
We will consider verifying NAS-inspired networks in the future work.

\bibliographystyle{splncs04}
\bibliography{egbib}

\begin{thebibliography}{10}
\providecommand{\url}[1]{\texttt{#1}}
\providecommand{\urlprefix}{URL }
\providecommand{\doi}[1]{https://doi.org/#1}

\bibitem{ahmed2018maskconnect}
Ahmed, K., Torresani, L.: Maskconnect: Connectivity learning by gradient
  descent. In: Proceedings of the European Conference on Computer Vision
  (ECCV). pp. 349--365 (2018)

\bibitem{bender2018understanding}
Bender, G., Kindermans, P.J., Zoph, B., Vasudevan, V., Le, Q.: Understanding
  and simplifying one-shot architecture search. In: International Conference on
  Machine Learning. pp. 550--559 (2018)

\bibitem{dai2017deformable}
Dai, J., Qi, H., Xiong, Y., Li, Y., Zhang, G., Hu, H., Wei, Y.: Deformable
  convolutional networks. In: Proceedings of the IEEE international conference
  on computer vision. pp. 764--773 (2017)

\bibitem{devries2017improved}
DeVries, T., Taylor, G.W.: Improved regularization of convolutional neural
  networks with cutout. arXiv preprint arXiv:1708.04552  (2017)

\bibitem{ghiasi2019fpn}
Ghiasi, G., Lin, T.Y., Le, Q.V.: Nas-fpn: Learning scalable feature pyramid
  architecture for object detection. In: Proceedings of the IEEE Conference on
  Computer Vision and Pattern Recognition. pp. 7036--7045 (2019)

\bibitem{girshick2018detectron}
Girshick, R., Radosavovic, I., Gkioxari, G., Doll{\'a}r, P., He, K.: Detectron
  (2018)

\bibitem{glorot2011deep}
Glorot, X., Bordes, A., Bengio, Y.: Deep sparse rectifier neural networks. In:
  Proceedings of the fourteenth international conference on artificial
  intelligence and statistics. pp. 315--323 (2011)

\bibitem{guo2019single}
Guo, Z., Zhang, X., Mu, H., Heng, W., Liu, Z., Wei, Y., Sun, J.: Single path
  one-shot neural architecture search with uniform sampling. arXiv preprint
  arXiv:1904.00420  (2019)

\bibitem{han2015learning}
Han, S., Pool, J., Tran, J., Dally, W.: Learning both weights and connections
  for efficient neural network. In: Advances in neural information processing
  systems. pp. 1135--1143 (2015)

\bibitem{he2016deep}
He, K., Zhang, X., Ren, S., Sun, J.: Deep residual learning for image
  recognition. In: Proceedings of the IEEE conference on computer vision and
  pattern recognition. pp. 770--778 (2016)

\bibitem{hinton2012improving}
Hinton, G.E., Srivastava, N., Krizhevsky, A., Sutskever, I., Salakhutdinov,
  R.R.: Improving neural networks by preventing co-adaptation of feature
  detectors. arXiv preprint arXiv:1207.0580  (2012)

\bibitem{howard2019searching}
Howard, A., Sandler, M., Chu, G., Chen, L.C., Chen, B., Tan, M., Wang, W., Zhu,
  Y., Pang, R., Vasudevan, V., et~al.: Searching for mobilenetv3. arXiv
  preprint arXiv:1905.02244  (2019)

\bibitem{howard2017mobilenets}
Howard, A.G., Zhu, M., Chen, B., Kalenichenko, D., Wang, W., Weyand, T.,
  Andreetto, M., Adam, H.: Mobilenets: Efficient convolutional neural networks
  for mobile vision applications. arXiv preprint arXiv:1704.04861  (2017)

\bibitem{huang2017densely}
Huang, G., Liu, Z., Van Der~Maaten, L., Weinberger, K.Q.: Densely connected
  convolutional networks. In: Proceedings of the IEEE conference on computer
  vision and pattern recognition. pp. 4700--4708 (2017)

\bibitem{huang2018data}
Huang, Z., Wang, N.: Data-driven sparse structure selection for deep neural
  networks. In: Proceedings of the European conference on computer vision
  (ECCV). pp. 304--320 (2018)

\bibitem{krizhevsky2009learning}
Krizhevsky, A., Hinton, G., et~al.: Learning multiple layers of features from
  tiny images. Tech. rep., Citeseer (2009)

\bibitem{krizhevsky2012imagenet}
Krizhevsky, A., Sutskever, I., Hinton, G.E.: Imagenet classification with deep
  convolutional neural networks. In: Advances in neural information processing
  systems. pp. 1097--1105 (2012)

\bibitem{lin2017feature}
Lin, T.Y., Doll{\'a}r, P., Girshick, R., He, K., Hariharan, B., Belongie, S.:
  Feature pyramid networks for object detection. In: Proceedings of the IEEE
  conference on computer vision and pattern recognition. pp. 2117--2125 (2017)

\bibitem{lin2014microsoft}
Lin, T.Y., Maire, M., Belongie, S., Hays, J., Perona, P., Ramanan, D.,
  Doll{\'a}r, P., Zitnick, C.L.: Microsoft coco: Common objects in context. In:
  European conference on computer vision. pp. 740--755. Springer (2014)

\bibitem{liu2019auto}
Liu, C., Chen, L.C., Schroff, F., Adam, H., Hua, W., Yuille, A.L., Fei-Fei, L.:
  Auto-deeplab: Hierarchical neural architecture search for semantic image
  segmentation. In: Proceedings of the IEEE Conference on Computer Vision and
  Pattern Recognition. pp. 82--92 (2019)

\bibitem{liu2018darts}
Liu, H., Simonyan, K., Yang, Y.: Darts: Differentiable architecture search.
  arXiv preprint arXiv:1806.09055  (2018)

\bibitem{loshchilov2016sgdr}
Loshchilov, I., Hutter, F.: Sgdr: Stochastic gradient descent with warm
  restarts. arXiv preprint arXiv:1608.03983  (2016)

\bibitem{luo2018differentiable}
Luo, P., Ren, J., Peng, Z., Zhang, R., Li, J.: Differentiable
  learning-to-normalize via switchable normalization. arXiv preprint
  arXiv:1806.10779  (2018)

\bibitem{maas2013rectifier}
Maas, A.L., Hannun, A.Y., Ng, A.Y.: Rectifier nonlinearities improve neural
  network acoustic models. In: Proc. icml. vol.~30, p.~3 (2013)

\bibitem{paszke2017automatic}
Paszke, A., Gross, S., Chintala, S., Chanan, G., Yang, E., DeVito, Z., Lin, Z.,
  Desmaison, A., Antiga, L., Lerer, A.: Automatic differentiation in pytorch
  (2017)

\bibitem{perez2018efficient}
P{\'e}rez-R{\'u}a, J.M., Baccouche, M., Pateux, S.: Efficient progressive
  neural architecture search. arXiv preprint arXiv:1808.00391  (2018)

\bibitem{ramachandran2017searching}
Ramachandran, P., Zoph, B., Le, Q.V.: Searching for activation functions. arXiv
  preprint arXiv:1710.05941  (2017)

\bibitem{rauschecker1984neuronal}
Rauschecker, J.: Neuronal mechanisms of developmental plasticity in the cat's
  visual system. Human neurobiology  (1984)

\bibitem{real2019regularized}
Real, E., Aggarwal, A., Huang, Y., Le, Q.V.: Regularized evolution for image
  classifier architecture search. In: Proceedings of the aaai conference on
  artificial intelligence. vol.~33, pp. 4780--4789 (2019)

\bibitem{russakovsky2015imagenet}
Russakovsky, O., Deng, J., Su, H., Krause, J., Satheesh, S., Ma, S., Huang, Z.,
  Karpathy, A., Khosla, A., Bernstein, M., et~al.: Imagenet large scale visual
  recognition challenge. International journal of computer vision
  \textbf{115}(3),  211--252 (2015)

\bibitem{sandler2018mobilenetv2}
Sandler, M., Howard, A., Zhu, M., Zhmoginov, A., Chen, L.C.: Mobilenetv2:
  Inverted residuals and linear bottlenecks. In: Proceedings of the IEEE
  conference on computer vision and pattern recognition. pp. 4510--4520 (2018)

\bibitem{simonyan2014very}
Simonyan, K., Zisserman, A.: Very deep convolutional networks for large-scale
  image recognition. arXiv preprint arXiv:1409.1556  (2014)

\bibitem{srivastava2015highway}
Srivastava, R.K., Greff, K., Schmidhuber, J.: Highway networks. arXiv preprint
  arXiv:1505.00387  (2015)

\bibitem{sun2019high}
Sun, K., Zhao, Y., Jiang, B., Cheng, T., Xiao, B., Liu, D., Mu, Y., Wang, X.,
  Liu, W., Wang, J.: High-resolution representations for labeling pixels and
  regions. arXiv preprint arXiv:1904.04514  (2019)

\bibitem{szegedy2017inception}
Szegedy, C., Ioffe, S., Vanhoucke, V., Alemi, A.A.: Inception-v4,
  inception-resnet and the impact of residual connections on learning. In:
  Thirty-first AAAI conference on artificial intelligence (2017)

\bibitem{szegedy2015going}
Szegedy, C., Liu, W., Jia, Y., Sermanet, P., Reed, S., Anguelov, D., Erhan, D.,
  Vanhoucke, V., Rabinovich, A.: Going deeper with convolutions. In:
  Proceedings of the IEEE conference on computer vision and pattern
  recognition. pp.~1--9 (2015)

\bibitem{tan2019mnasnet}
Tan, M., Chen, B., Pang, R., Vasudevan, V., Sandler, M., Howard, A., Le, Q.V.:
  Mnasnet: Platform-aware neural architecture search for mobile. In:
  Proceedings of the IEEE Conference on Computer Vision and Pattern
  Recognition. pp. 2820--2828 (2019)

\bibitem{tan2019efficientnet}
Tan, M., Le, Q.V.: Efficientnet: Rethinking model scaling for convolutional
  neural networks. arXiv preprint arXiv:1905.11946  (2019)

\bibitem{tang2018quantized}
Tang, Z., Peng, X., Geng, S., Wu, L., Zhang, S., Metaxas, D.: Quantized densely
  connected u-nets for efficient landmark localization. In: Proceedings of the
  European Conference on Computer Vision (ECCV). pp. 339--354 (2018)

\bibitem{DBLP:conf/nips/VeitWB16}
Veit, A., Wilber, M.J., Belongie, S.J.: Residual networks behave like ensembles
  of relatively shallow networks. In: {NIPS}. pp. 550--558 (2016)

\bibitem{venables2013modern}
Venables, W.N., Ripley, B.D.: Modern applied statistics with S-PLUS. Springer
  Science \& Business Media (2013)

\bibitem{weinberger2009feature}
Weinberger, K., Dasgupta, A., Langford, J., Smola, A., Attenberg, J.: Feature
  hashing for large scale multitask learning. In: Proceedings of the 26th
  annual international conference on machine learning. pp. 1113--1120 (2009)

\bibitem{wu2018group}
Wu, Y., He, K.: Group normalization. In: Proceedings of the European Conference
  on Computer Vision (ECCV). pp. 3--19 (2018)

\bibitem{xie2019exploring}
Xie, S., Kirillov, A., Girshick, R., He, K.: Exploring randomly wired neural
  networks for image recognition. arXiv preprint arXiv:1904.01569  (2019)

\bibitem{zhang2018shufflenet}
Zhang, X., Zhou, X., Lin, M., Sun, J.: Shufflenet: An extremely efficient
  convolutional neural network for mobile devices. In: Proceedings of the IEEE
  conference on computer vision and pattern recognition. pp. 6848--6856 (2018)

\bibitem{zoph2018learning}
Zoph, B., Vasudevan, V., Shlens, J., Le, Q.V.: Learning transferable
  architectures for scalable image recognition. In: Proceedings of the IEEE
  conference on computer vision and pattern recognition. pp. 8697--8710 (2018)

\end{thebibliography}
\end{document}